\documentclass[pdflatex,sn-mathphys-num]{sn-jnl}

\usepackage{graphicx}%
\usepackage{pifont}
\usepackage{subcaption} 
\usepackage{multirow}%
\usepackage{amsmath,amssymb,amsfonts}%
\usepackage{amsthm}%
\usepackage{mathrsfs}%
\usepackage[title]{appendix}%
\usepackage{xcolor}%
\usepackage{textcomp}%
\usepackage{manyfoot}%
\usepackage{booktabs}%
\usepackage{algorithm}%
\usepackage{algorithmicx}%
\usepackage{algpseudocode}%
\usepackage{array}
\usepackage{booktabs}
\usepackage{siunitx}
\usepackage{listings}%
\usepackage[T1]{fontenc}
\usepackage{anyfontsize}
\DeclareFontFamily{U}{rsfs}{\skewchar\font127 }
\DeclareFontShape{U}{rsfs}{m}{n}{ <-> s*[.9] rsfs10 }{}

\usepackage{listings}
\theoremstyle{thmstyleone}%
\theoremstyle{thmstyletwo}%

\theoremstyle{thmstylethree}%

\begin{document}

\title[Article Title]{DrugPilot: LLM-based Parameterized Reasoning Agent for Drug Discovery}







\author[1]{\fnm{Kun} \sur{Li}}\email{likun98@whu.edu.cn}
\equalcont{These authors contributed equally to this work.}

\author[1]{\fnm{Zhennan} \sur{Wu}}\email{wuzhennan@whu.edu.cn}
\equalcont{These authors contributed equally to this work.}

\author[2]{\fnm{Shoupeng} \sur{Wang}}\email{wangshoupeng@whu.edu.cn}
\equalcont{These authors contributed equally to this work.}

\author[3]{\fnm{Jia} \sur{Wu}}\email{jia.wu@mq.edu.au}
 
\author[4]{\fnm{Shirui} \sur{Pan}}\email{s.pan@griffith.edu.au}

\author*[1]{\fnm{Wenbin} \sur{Hu}}\email{hwb@whu.edu.cn}

\affil[1]{\orgdiv{School of Computer Science}, \orgname{Wuhan University}, \city{Wuhan}, \country{China}}
\affil[2]{\orgdiv{School of Mathematics and Statistics}, \orgname{Wuhan University}, \city{Wuhan}, \country{China}}
\affil[3]{\orgdiv{Department of Computing}, \orgname{Macquarie University}, \city{Sydney}, \country{Australia}}
\affil[4]{\orgdiv{School of Information and Communication Technology}, \orgname{Griffith University}, \city{Brisbane}, \country{Australia}}


\abstract{

Large language models (LLMs) integrated with autonomous agents hold significant potential for advancing scientific discovery through automated reasoning and task execution. However, applying LLM agents to drug discovery is still constrained by challenges such as large-scale multimodal data processing, limited task automation, and poor support for domain-specific tools. To overcome these limitations, we introduce \textbf{DrugPilot}, a LLM-based agent system with a parameterized reasoning architecture designed for end-to-end scientific workflows in drug discovery. DrugPilot enables multi-stage research processes by integrating structured tool use with a novel parameterized memory pool. The memory pool converts heterogeneous data from both public sources and user-defined inputs into standardized representations. This design supports efficient multi-turn dialogue, reduces information loss during data exchange, and enhances complex scientific decision-making. To support training and benchmarking, we construct a drug instruction dataset covering eight core drug discovery tasks. Under the Berkeley function-calling benchmark, DrugPilot significantly outperforms state-of-the-art agents such as ReAct and LoT, achieving task completion rates of 98.0\%, 93.5\%, and 64.0\% for simple, multi-tool, and multi-turn scenarios, respectively. These results highlight  DrugPilot's potential as a versatile agent framework for computational science domains requiring automated, interactive, and data-integrated reasoning.

}

\keywords{large language model, agent, tool calling, parameterized reasoning, drug discovery}



\maketitle


With the rapid development of deep learning, artificial intelligence (AI)-assisted drug discovery is emerging as a revolutionary way to significantly enhance the efficiency and accuracy of key phases of complex drug discovery tasks \cite{aidd1,aidd2}. For example, Chemistry42 \cite{Chemistry42}, a generative AI tool, MolProphet \cite{MolProphet}, a drug screening system, and DrugFlow \cite{drugflow}, an AI platform, have demonstrated excellent prediction performance and dramatically shortened the drug development period. Recent studies, based on these, have applied large language models (LLMs) to automating drug discovery tasks \cite{llmdd1,llmdd2,llmdd3}. The DrugAgent \cite{drugagent2}, for example, demonstrated the potential of LLMs for drug R\&D automation by setting up two intelligent body roles, the mentor and the planner, and constructing an end-to-end machine learning process from data acquisition to model evaluation. Another study with the same name \cite{drugagent1} developed an agent system for drug repositioning by combining knowledge graphs and literature mining techniques. These Agents can leverage the reasoning capabilities of LLMs and perform well in handling multi-source heterogeneous data, demonstrating that LLMs not only facilitate cross-domain knowledge fusion, but also hold the promise of automating the whole process of drug discovery through intelligent decision-making \cite{Zheng2025}.

\begin{figure*}[t!]
    \centering
    \includegraphics[width=1\linewidth]{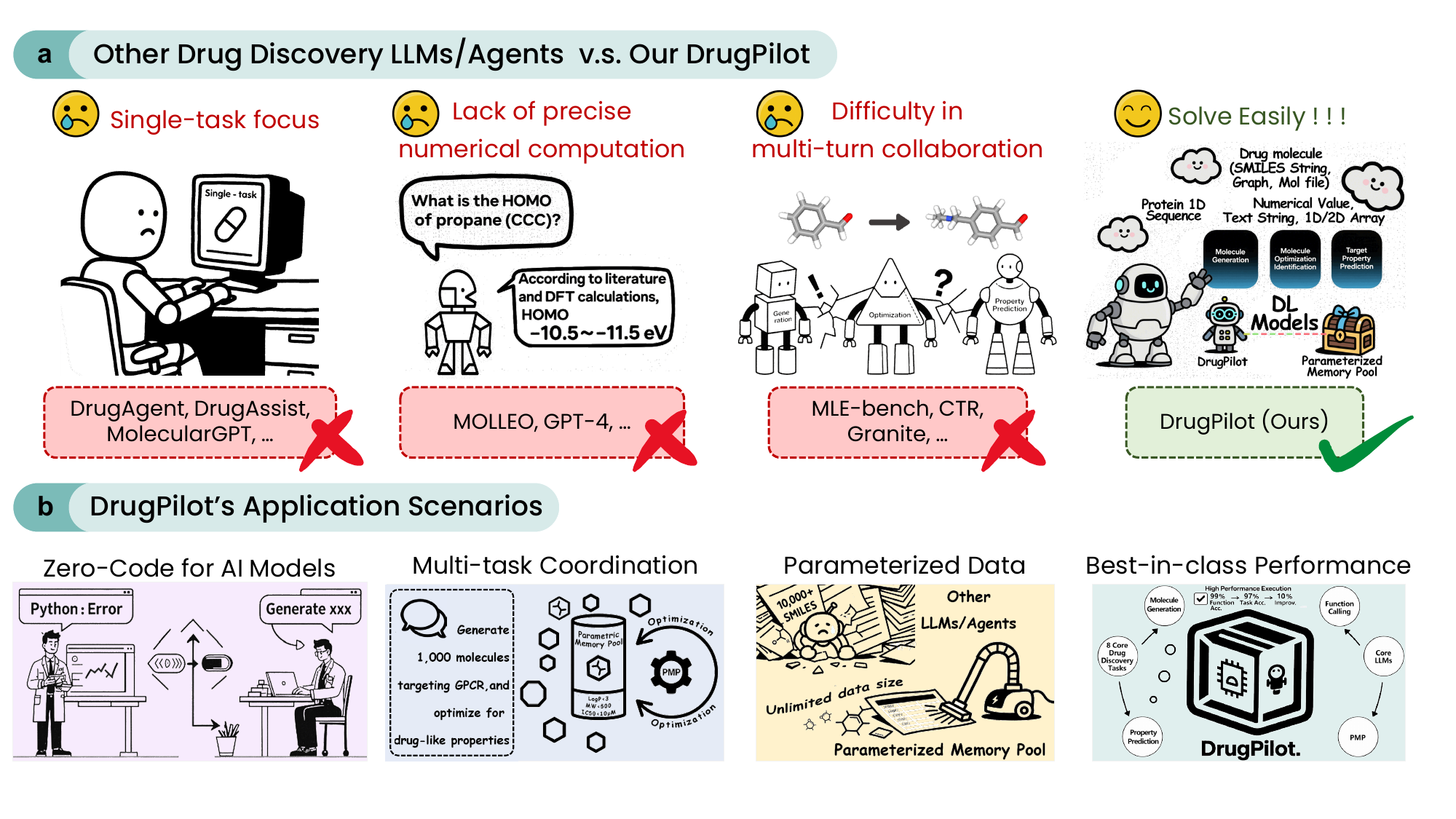}
    \caption{\textbf{Application scenarios and advantages of DrugPilot. a,} Four application scenarios of DrugPilot: zero-code integration of AI models, scalable data acquisition, coordinated multi-task processing, and accurate execution of 8 essential drug discovery tasks. \textbf{b,} Comparison of LLMs or agents for drug discovery with our DrugPilot across three aspects.}
    \label{fig:mov}
\end{figure*}

Although fine-tuning and reinforcement learning can effectively improve the task performance of LLMs, leveraging their reasoning capabilities for automated drug discovery and accurate prediction still faces significant challenges (Fig. \ref{fig:mov}a) \cite{de2023transforming,tiwari2023artificial}.  The primary challenge is to offer convenient usage for non-computer science users.  Drug discovery is a multi-stage, time-consuming process involving a series of complex tasks \cite{complex}, including drug generation and optimization  \cite{molopt}, target affinity prediction \cite{dta1}, and molecular property prediction  \cite{cmk,mevon} and so on.  However, researchers in pharmacy, biology, and other related fields often lack the technical expertise to operate state-of-the-art (SOTA) deep learning models \cite{sadybekov2023computational,mak2019artificial}.  These models have stringent input format requirements, and researchers need to spend a lot of time pre-processing real-world data and converting data formats between different platforms \cite{niazi2023computer,chen2021applications}.  Lack of knowledge of programming, computer hardware, and system operation significantly increases the cost for researchers to use SOTA models effectively, and in some cases hinders their use completely \cite{zhavoronkov2019deep,stokes2020deep}.


In addition, there are various challenges in designing LLM-based agents, such as multi-task collaboration, large-scale multimodal data management, and predictive performance requirements during complex task computation.  Most existing platforms operate as standalone tools and lack automated task planning and execution capabilities \cite{drugagent2,drugagent1,ren2025small}, while drug discovery usually involves more than one phase and requires the seamless integration between tasks \cite{zhang2025artificial}.  This limitation forces researchers to manually switch between tools and integrate intermediate outputs, significantly reducing the efficiency of human–computer collaboration \cite{llmdd1,llmdd3}.  More critically, the natural language output format of LLMs inherently restricts their ability to precisely represent domain-specific entities (e.g., drug compounds, cell lines, and targets) or to reliably predict their interactions.  This often results in hallucinations and fragmented multimodal reasoning \cite{llmdd1,llmdd3, Zheng2025}, as evidenced by failed semantic alignment across molecular graphs, bioactivity data (numerical matrices), and literature-based evidence (natural language).  As data volume increases, the text-based memory paradigm of LLMs tends to lose critical information, leading to task interruptions and failures.  Moreover, currently, LLMs perform poorly when dealing with complex tasks, particularly in terms of tool calling accuracy and multi-turn conversation capabilities. As a result, users are unable to carry out sustained practical research. These real-world challenges substantially hinder the ability of LLM-based agents to complete tasks accurately and efficiently (Fig. \ref{fig:mov}a).


\begin{figure}[t!]
    \centering
    \includegraphics[width=1\linewidth]{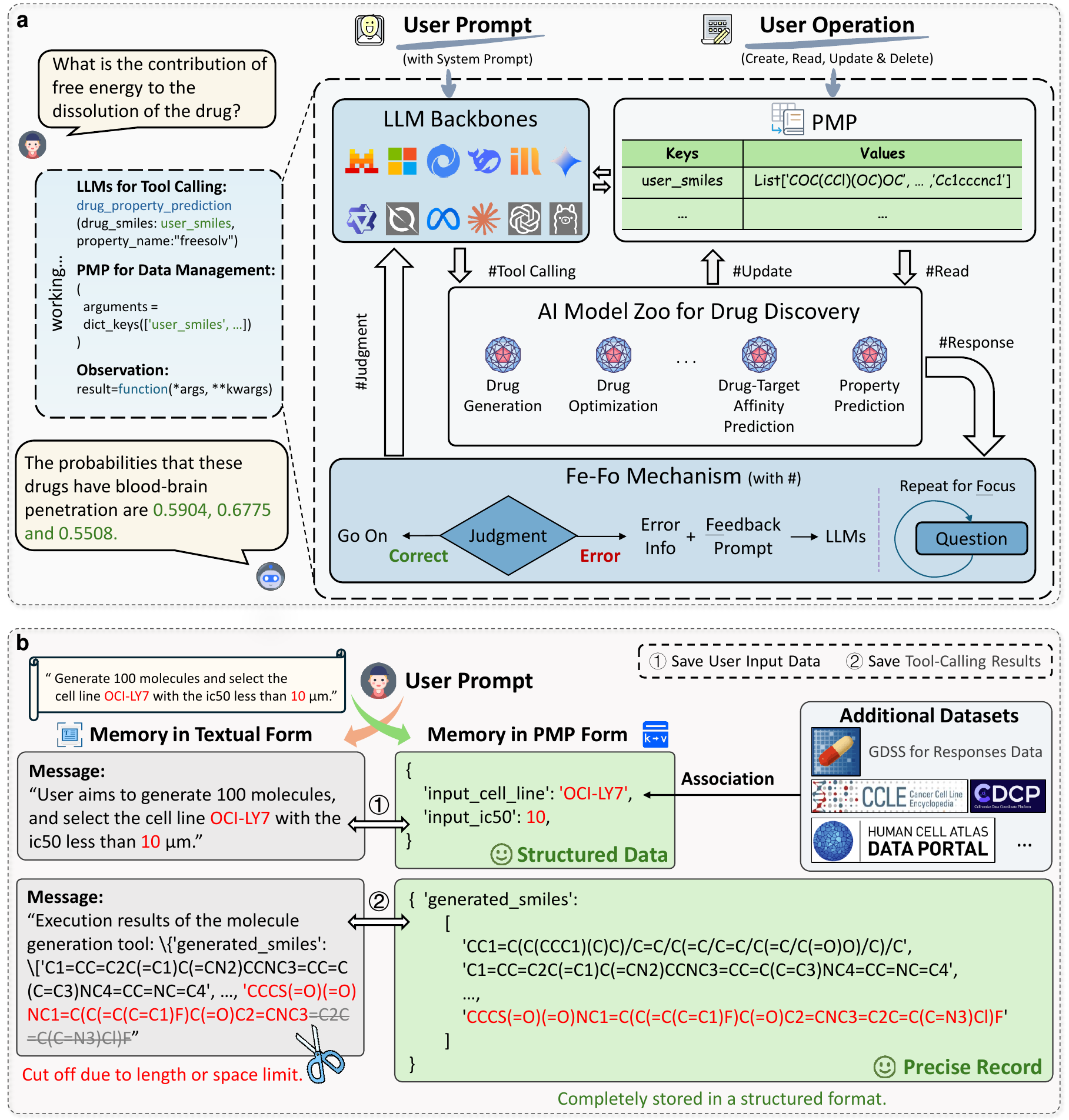}
    \caption{\textbf{Architecture of DrugPilot framework. a,} The framework of DrugPilot. DrugPilot comprises four components: the LLM backbones, the PMP, the AI models tailored to 8 stages of drug discovery with tool calling support, and the Fe-Fo mechanism. \textbf{b,} The PMP's structure. In PMP, the memory is stored as key-value pairs, where LLMs interact only with concise keys, while tools directly interact with the structured values.}
    \label{fig:framework}
\end{figure}

To address these challenges, we propose DrugPilot, an LLM-based agent with parameterized reasoning for drug discovery (Fig. \ref{fig:framework}). DrugPilot comprehensively supports the entire drug research and development pipeline and can autonomously plan and execute multi-stage research tasks based on user queries (Fig. \ref{fig:mov}b). To meet the critical demands for accurate extraction and analysis of multimodal drug data, including both public datasets and user-provided inputs, we propose an interactive parameterized memory pool (PMP). As a highly flexible component, PMP converts real-world drug data into standardized parametric representations. This design enables efficient knowledge retrieval during multi-turn interactions and mitigates the information loss commonly associated with text-based transmission. To overcome common reasoning errors encountered by LLMs when interpreting PMP and invoking tools, as well as their tendency to lose track of the original task in extended dialogues, we further introduce a feedback-focus mechanism, called Fe-Fo. Meanwhile, We propose the first tool-calling benchmark for drug discovery, which includes a high-quality instruction dataset, TCDD, and an evaluation method. TCDD consists of 2,800 annotated samples spanning 8 representative drug discovery tasks. On this benchmark, DrugPilot demonstrates superior performance compared to existing approaches including ReAct \cite{react}, CoT \cite{cot}, and Lot \cite{lot}, as evaluated against the Berkeley function-calling leaderboard. Specifically, DrugPilot achieves task completion rates of 98.0\%, 93.5\%, and 64.0\% across three evaluation categories, increasing by 13.2\%, 66.1\%, and 80.3\%, respectively, over the SOTA agent ReAct.

\section*{Results}\label{sec2}

\subsection*{DrugPilot's Framework}


We propose DrugPilot, an LLM-based agent with parameterized reasoning for drug discovery. DrugPilot comprehensively supports the entire drug research and development pipeline, which can autonomously plan and execute multi-stage research tasks based on user queries. The DrugPilot system comprises four key components: the LLMs, the PMP, the Fe-Fo mechanism, and the AI model zoo. DrugPilot operates through a collaborative framework combining natural language interaction and the PMP (Fig. \ref{fig:framework}). The PMP is proposed for extracting textual information and processing large-scale multimodal data, and Fe-Fo mechanism could perform real-time monitoring of the LLMs' outputs. Fe-Fo targets errors that occur when LLMs read PMP and call tools, providing specific error feedback to help LLMs correct their mistakes, and restating the original question to help LLMs maintain focus.  The AI model zoo encompasses a collection of state-of-the-art deep learning models that support core tasks in drug discovery (see Fig. \ref{fig:sharegpt}e for task details). This paper focuses on the design of LLM-based agents to enable more efficient tool invocation and task-data coordination. Each task-specific model can be seamlessly integrated into the DrugPilot framework. Additional implementation details and performance benchmarks for task-specific models are available in the Supplementary Materials.


DrugPilot enables users to complete preclinical drug research in real time through natural language conversations. In addition, DrugPilot supports large-scale file uploads, automatically parses and saves file data to the PMP, and the parameters and prediction results during the dialogue process are also saved to the PMP, supporting operations such as downloading, modifying, and deleting. The DrugPilot delivers outputs in two formats: providing conclusions through natural language dialogue while presenting data visualizations via the PMP. For example, in molecular optimization tasks, user's input is:
\begin{lstlisting}
Please generate some optimized molecules based on the molecule
CC(C)C1=CC=CC=C1CC2=C(C(=C(C(=C2)C(=O)NC3=CC=C(C=C3)
S(=O)(=O)C4=CC=CC=C4C(C)(C)C)O)O)O,
the Z-score of cell line MOLT-13 is required to be less than -0.47.
\end{lstlisting}
DrugPilot first interprets the user's input requirements, identifies the task type and key parameters, including the molecules to be optimized and the optimization conditions, and then stores this information in the PMP. DrugPilot then invokes the conditional molecular optimization model and passes the parameters from the PMP to the model. Once DrugPilot completes execution, it generates a specified number of optimized molecules. These molecules have been stored in the PMP, enabling users to download them or proceed to in-depth analysis. Therefore, this parameterized data management mode accurately records molecular SMILES, enabling seamless integration with other tasks such as molecular property prediction. Compared to traditional LLM-based methods, DrugPilot significantly advances molecular optimization by generating hundreds to thousands of precise SMILES candidates per cycle (vs. dozens) and dynamically integrating SOTA algorithms, overcoming the limitations of fixed AI models and stagnant SOTA performance.



\subsection*{Superior Performance over SOTA LLMs and Agents}
\label{overallexp}

To comprehensively assess the performance of DrugPilot, we conducted an overall experiment assessing combinations of different LLMs and various agent paradigms. These agent paradigms serve as baseline methods for comparison with DrugPilot, while the different LLMs are used to examine the overall performance of these methods when paired with different models. DrugPilot can be directly used with pretrained LLMs; however, to further enhance its performance, we also fine-tuned the pretrained LLMs specifically for scenarios involving drug-related tool usage. In this experiment, we treat using DrugPilot with pretrained LLMs and with fine-tuned LLMs as two separate methods for comparison.



Table~\ref{tab:model_performance} reports the experimental results. DrugPilot achieves the highest accuracy on the three categories over all agents. For the simple function category, mainstream LLMs including Llama3.1 \cite{llama3_1}, Llama3 \cite{llama3}, Mistral-NeMo \cite{mistral}, Gemma2 \cite{team2024gemma}, and Qwen2 \cite{qwen2} all exceed 97\% on Acc.F and 95\% on Acc.P. 
In the multi-function category, DrugPilot continues to dominate. Across Llama3.1, Llama3, Mistral-NeMo, and Gemma2, Acc.F stays above 98\% and Acc.P stays above 92\%.
Even in the most complex category, multi-turn function, DrugPilot achieves superior performance, attaining accuracy rates exceeding 70\% and 60\% compared to the SOTA methods.

As the difficulty of tool calling scenarios increases, the accuracy declines overall. But on all three categories, DrugPilot outperforms the SOTA agent ReAct by 13.6\% and 13.2\% in simple function, 29.4\% and 66.1\% in multi-function, 61.2\% and 80.3\% in multi-turn function on Acc.F and Acc.P. Notably, though less capable LLM like deepseek-llm-7b-chat performs poor, especially in the category of multi-turn function, it still gains improvement compared to baseline methods. Besides, after removing SFT, the performance of DrugPilot declines obviously, but it still surpasses the baseline methods in most scenarios. Compared with the second-best method, DrugPilot still has considerable improvement, with 6.3\% and 10.9\% in simple function, 24.3\% and 60.6\% in multi-function, 46.9\% and 67.5\% in multi-turn function on Acc.F and Acc.P.

\begin{figure*}[t!]
    \centering
    \includegraphics[width=1\linewidth]{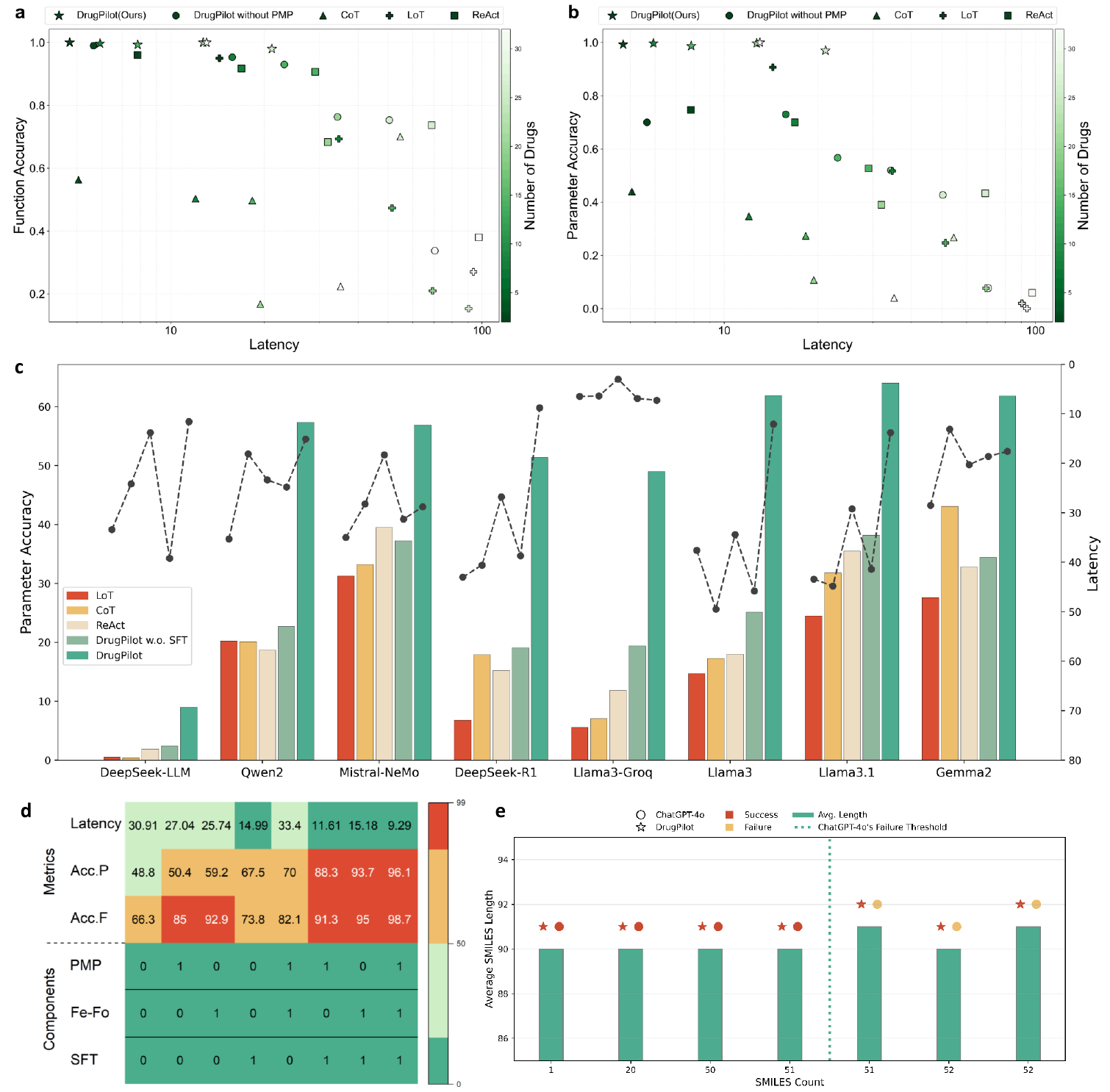}
    \caption{\textbf{Visualization of the experimental results. a,} Function accuracy of different agent methods under varying molecule quantities. \textbf{b,} Parameter accuracy of different agent methods under varying molecule quantities. \textbf{c,} The parameter accuracy and latency of LLMs and agents on category multi-turn function. \textbf{d,} Ablation studies about the effects of SFT, Fe-Fo, and PMP components on multi-task performance metrics in DrugPilot’s Llama3.1-8B framework. \textbf{e,} Evaluating the capacity boundaries of ChatGPT-4o and DrugPilot for drug discovery tasks.}
    \label{fig:ablation_mp}
\end{figure*}

In addition to measuring accuracy, we also recorded the average execution time for each query in the multi-turn function category. As shown in Fig.\ref{fig:ablation_mp}c, the average latency of DrugPilot has been reduced to under 20s, and for models including Qwen2, DeepSeek-R1, Llama3, and Llama3.1, this means more than a twofold improvement over baseline methods. The decrease is particularly striking given that the multi-turn tasks require sequential reasoning based on former results. In practice, this means that DrugPilot not only delivers substantially 
higher parameter accuracy but does so with a runtime that is less than half of what SOTA agents require. The combination of SFT and an optimized tool‐calling procedure allows DrugPilot to maintain fast response times even as task complexity grows, demonstrating that accuracy and efficiency gains can be achieved simultaneously.



\subsection*{Parameterized Memory Pool enables High-volume Drug Data Processing}

The excessive scale of parameters is a key challenge in utilizing LLM-based agents for drug discovery tasks. To evaluate the upper limit of the parameter scale that existing methods can handle when processing drug-related parameters, we conducted a parameter scale experiment. We tested ChatGPT-4o \cite{gpt4o}, one of the most powerful closed-source large-scale LLMs, in a drug tool-calling scenario. We compared its performance with DrugPilot by invoking ChatGPT-4o's tool-calling API. The evaluation task involved calling a drug property prediction tool to predict the water solubility of molecules. By adjusting the number of drug molecules and their average string length in the input, we investigated its capability to handle large-scale drug parameters.  

The capacity boundaries of LLMs are shown in Fig.~\ref{fig:ablation_mp}e. When the number of input molecules was $\leq$ 51 and the average molecular length was $\leq$ 90, ChatGPT-4o could stably select the correct tool, accurately pass the parameters, and ensure successful task execution. However, when these thresholds were exceeded, constrained by the model's context length, ChatGPT-4o failed to extract and output such large-scale parameters. These results indicate that ChatGPT-4o's performance in drug property prediction tasks is limited by the scale of input drug molecules, which must be considered in practical applications involving large-scale parameters in drug discovery.  

In contrast, DrugPilot has no upper limit on parameter scale. Even with 91 molecules and an average length of 52, it could still accurately pass the parameters to the tool. Since the PMP employs a key-value pair conversion mechanism to remove large-scale parameters from the LLM's context, DrugPilot can theoretically handle parameters of any scale. This effectively solves the critical issue of large-scale parameter transfer in drug discovery agents.  



\begin{figure*}[t!]
    \centering
    \includegraphics[width=1\linewidth]{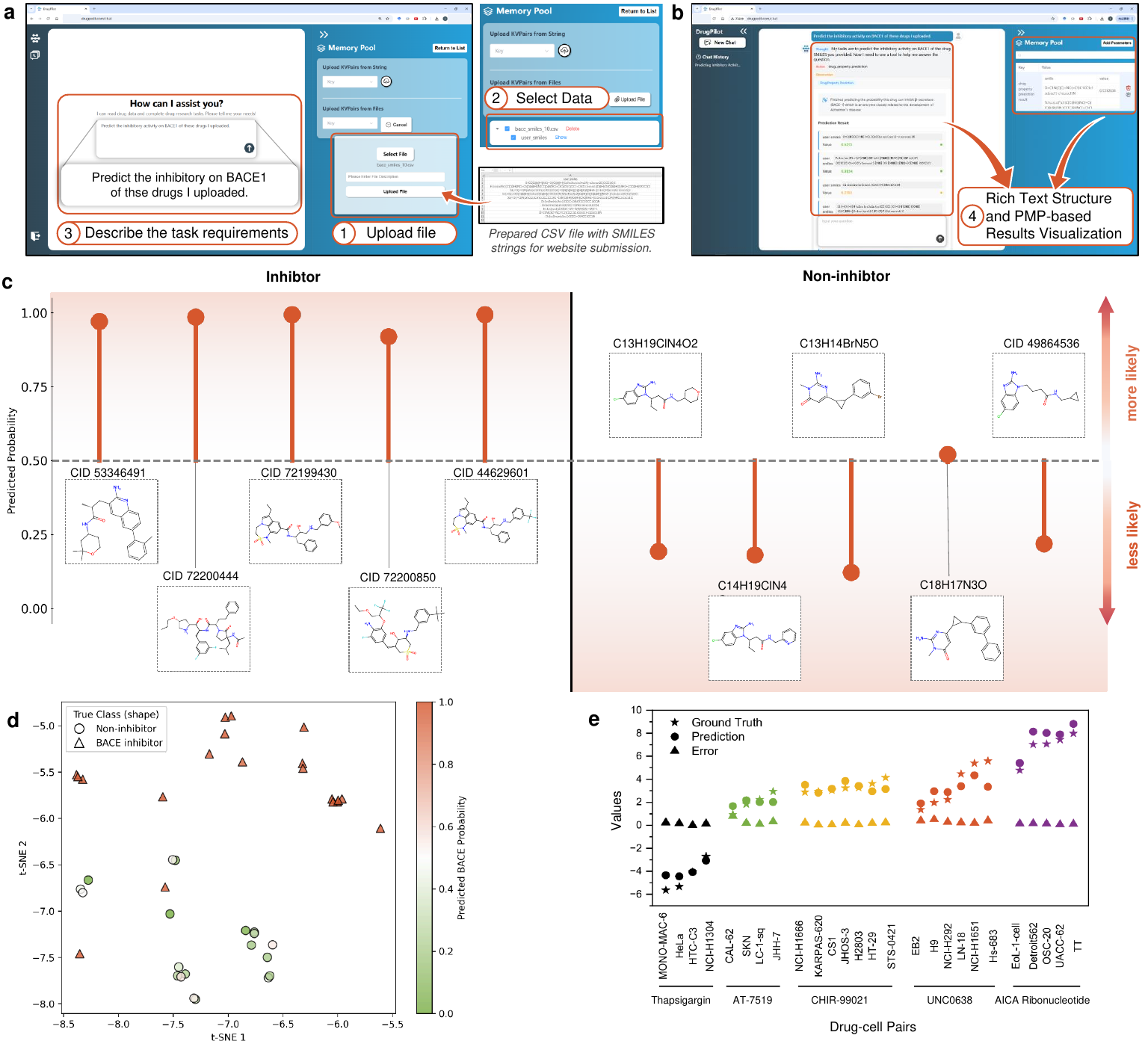}
    \caption{\textbf{Case study. a-b,} A case of using the DrugPilot platform to predict BACEi classes. \textbf{a,} Data upload process on DrugPilot.
    \textbf{b,} Prediction result display of DrugPilot. \textbf{c-d} Prediction results of DrugPilot on BACEi classification task. \textbf{c,} Predicted probabilities for 10 SMILES as BACEi. The compound IDs (CIDs) of known molecules and molecular formula (MF) expressions of unknown molecules are shown. \textbf{d,} Distribution of BACEi molecules in the t-SNE space. \textbf{e,} Comparative analysis of cross-cell-line drug efficacy prediction and experimental validation for five classes of anticancer drugs.}
    \label{fig:case_study}
\end{figure*}

\subsection*{Case Studies on Two Classical Drug Discovery Tasks}

This section presents the systematic case studies to evaluate the actual effect of the DrugPilot platform on drug discovery tasks. We select two core tasks as evaluation targets: drug response prediction (DRP) and molecular property prediction (MPP). The successful application of DrugPilot relies on two fundamental capabilities: (1) accurately selecting functional modules and extracting task-specific parameters; and (2) correctly invoking embedded models to produce reliable predictions. As demonstrated in Section~\ref{overallexp}, the platform's capability 1 has been thoroughly verified. In this study, we construct a testing framework on the basis of the Genomics of Drug Sensitivity in Cancer (GDSC) v2 \cite{gdsc} and BACE \cite{bace} dataset, simulating realistic drug development scenarios to assess the platform's predictive accuracy, specifically its capability 2.

\paragraph{Model Capabilities for the DRP Task.} The DRP task evaluates drug sensitivity or efficacy for specific cell lines using quantitative metrics such as IC50, AUC, or cell viability. The model takes drug molecular features (e.g., molecular structures or fingerprints) and cell line characteristics (e.g., gene expression profiles or mutation information) as input, and outputs continuous predictive values representing drug response effects. 

In this case study, the DrugPilot platform employs the CLDR method \cite{CLDR} as its core predictive engine. As illustrated in Fig. \ref{fig:case_study}, the DrugPilot platform provides a comprehensive DRP implementation solution. Users can submit drug-cell line pairs for analysis through an intuitive file upload interface. This study utilizes test datasets from the GDSCv2, with careful selection of five drug-cell line combinations (4, 4, 7, 6, and 5 samples per group, respectively, totaling 26 test samples) that were excluded from model training for validation purposes. Fig. \ref{fig:case_study}c presents the performance evaluation results of the DrugPilot platform. Analytical results demonstrate that across five independent drug-cell line tests, the platform's predicted values exhibit highly consistent monotonically increasing trends when aligned with the ascending order of actual measurements. Notably, the stable distribution of prediction errors around zero robustly validates the superior precision and reliability of the DrugPilot platform.

\paragraph{Model Capabilities for the MPP Task.}
The MPP task aims to infer essential molecular-level characteristics. Depending on the type of property, the model outputs either classification probabilities or regression values. MPP plays a critical role in early-stage drug screening and compound prioritization.

DrugPilot employs KCHML \cite{KCHML} as the core engine for MPP. In this case study, we evaluate its predictive capabilities on a binary classification task that determines whether a molecule functions as a BACE inhibitor. As shown in Fig.~\ref{fig:case_study}c, we select the top five and bottom five molecules from the BACE dataset and predict their likelihood of being BACE-1 inhibitors (BACEi) using DrugPilot. The results show that most predicted probabilities are highly consistent with the ground truth labels. Fig.~\ref{fig:case_study}d further demonstrates the classification performance by visualizing the top 20 and bottom 20 molecules from the BACE dataset. We apply t-SNE to project their fingerprint features into a two-dimensional space. The resulting distribution reveals a strong alignment between the true BACEi and those predicted BACEi with high probability, indicating the model’s ability to effectively distinguish molecular classes. Together, these findings validate the high accuracy and reliability of DrugPilot in molecular property prediction tasks.

\subsection*{Study of key components}
To comprehensively evaluate the impact of each component within DrugPilot on overall system performance, we conducted ablation studies from three perspectives: SFT, the feedback-focus (Fe-Fo) mechanism, and the PMP (Parameterized Memory Pool). These experiments aim to analyze the individual contribution and interplay of each module. All experiments used Llama3.1-8B as the foundation LLM under a simplified tool-calling scenario.


\paragraph{Effect of SFT and Fe-Fo Mechanism.} To investigate the role of SFT and Fe-Fo, we removed each module separately while keeping all other conditions unchanged. We then evaluated the accuracy of tool selection and parameter extraction across multi-functional tasks. As shown in  Fig.~\ref{fig:ablation_mp}d, both modules have a significant positive effect on performance. 
With the addition of SFT and Fe-Fo, tool-selection accuracy achieves 95.0\% and parameter-extraction accuracy achieves 93.7\%, increasing by 28.7\% and 44.9\% respectively. Notably, the latency has also been greatly improved, from the original 30.91s to 15.18s. Compared to the baseline results in Table.~\ref{tab:model_performance}, either component results in a notable accuracy improvement. 
In comparison with the baseline method, the addition of SFT and Fe-Fo increase tool‐selection accuracy to 73.8\% (+7.5\%) and 82.1\% (+15.8\%), parameter‐extraction accuracy to 59.2\% (+10.4\%) and 70.0\% (+21.2\%). Furthermore, when PMP is enabled, the performance improvement brought by SFT is greater, with parameter‐extraction accuracy to 88.3\% (+37.9\%). Meanwhile, when the model undergoes SFT, Fe-Fo also increases the parameter‐extraction accuracy more significantly, reaching 93.7\% (+26.2\%).
Specifically, SFT effectively enhances the model’s understanding of domain-specific text, while the Fe-Fo mechanism significantly improves the agent’s error handling ability.


\paragraph{Effect of PMP.} This study highlights the introduction of PMP, designed to handle large-scale molecular data. To validate the effectiveness of PMP, we gradually increased the number of molecules processed per query and tested the boundary of parameter recognition and extraction capability for different agent methods. To simulate real-world usage, an equivalent number of molecules were loaded into the memory pool. Under the same test data and model settings, we recorded the accuracy and latency of various agents. Fig.~\ref{fig:ablation_mp}a and Fig.~\ref{fig:ablation_mp}b show that the performance of traditional approaches degrades as the number of molecules increases—accuracy drops continuously while latency rises sharply. In particular, when the number of molecules exceeds 15, other methods' accuracy drops significantly. When the number of molecules reaches 20, the CoT method essentially fails to extract valid parameters. Similarly, DrugPilot without PMP, as well as ReAct and LoT, become nonfunctional when handling around 30 molecules. In contrast, real-world applications often require processing tens of thousands of molecules at once. DrugPilot equipped with PMP, however, maintains stable accuracy and response time throughout. The incorporation of PMP enables the model to focus on understanding user intent, effectively overcoming the challenge of extracting task-relevant parameters from lengthy and complex molecular descriptions.

\begin{figure*}[t!]
    \centering
    \includegraphics[width=\linewidth]{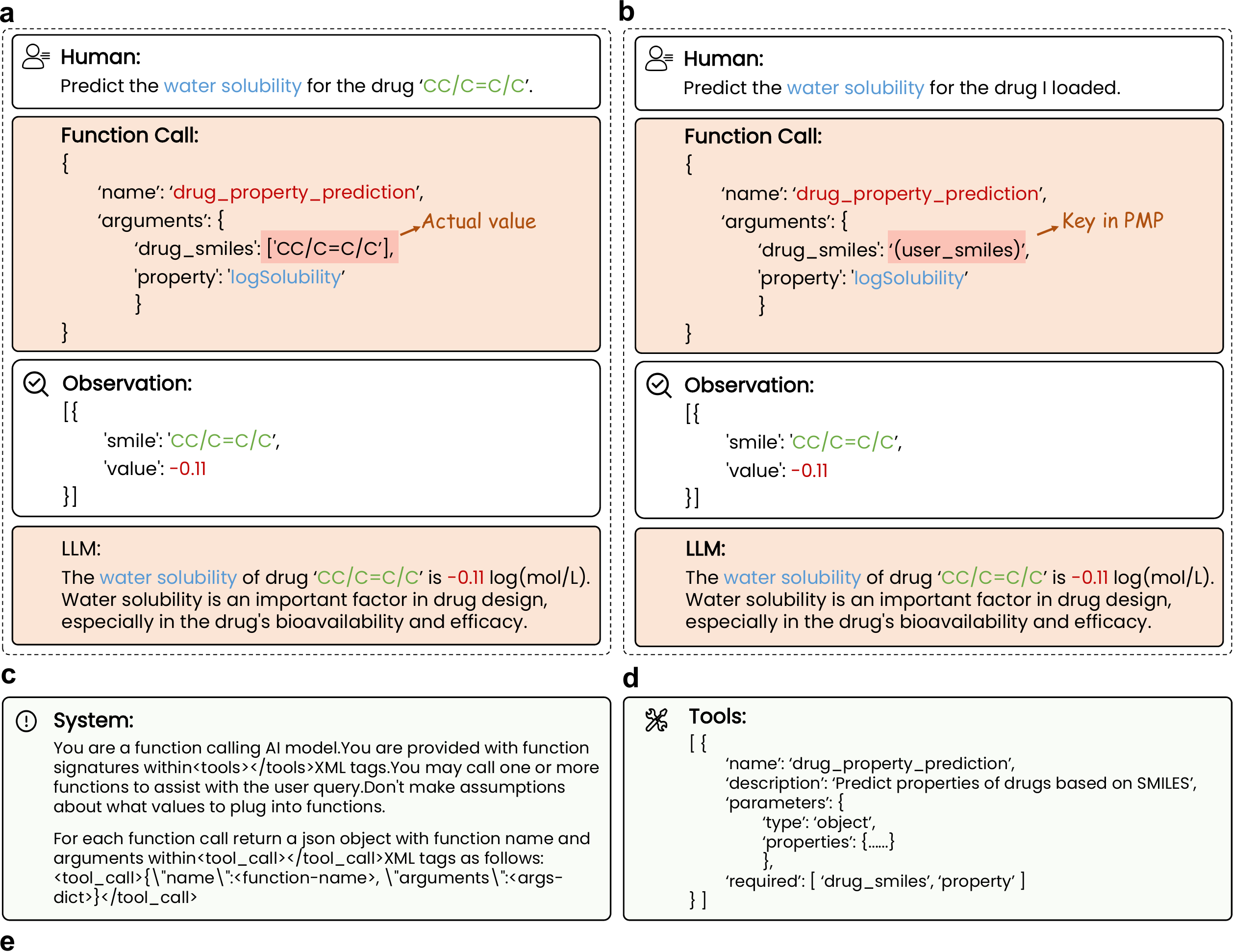}
    \includegraphics[width=\linewidth]{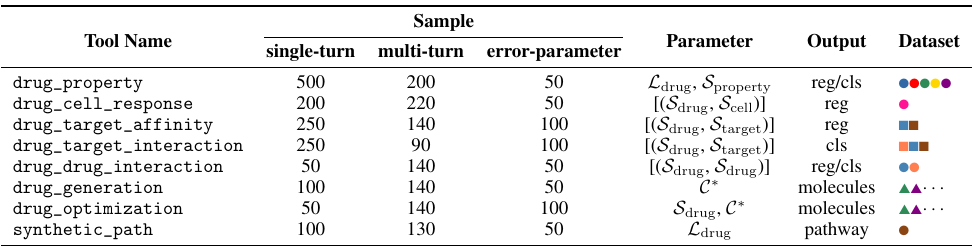}
    \begin{minipage}{\textwidth} 
    \tiny
    \textbf{Notations:}\\
    \textcircled{\raisebox{-0.9pt}{1}} There are different formalized parameters corresponding to keys in the memory pool, this sample takes \texttt{(user\_smiles)} for an example.\\
    \textcircled{\raisebox{-0.9pt}{2}} \textit{Function Call} is the proper key in ShareGPT format, and we identify function calling and tool calling in this paper.\\
    \textcircled{\raisebox{-0.9pt}{3}} The capital letters denote the types of corresponding parameters, with $\mathcal{L}$ standing for list and $\mathcal{S}$ standing for string. Tools \texttt{drug\_generation} and \texttt{drug\_optimization} integrate multiple models trained on different datasets, and they are suitable for different conditions. We represent their parameters with $\mathcal{C}^*$ uniformly.\\
    \textcircled{\raisebox{-0.9pt}{4}} Dataset: 
    \textcolor[HTML]{3069a5}{\ding{108}}~BACE, 
    \textcolor[HTML]{FF0000}{\ding{108}}~BBBP, 
    \textcolor[HTML]{2E8B57}{\ding{108}}~ESOL, 
    \textcolor[HTML]{FFD700}{\ding{108}}~FreeSolv, 
    \textcolor[HTML]{800080}{\ding{108}}~LIPO, 
    \textcolor[HTML]{FF1493}{\ding{108}}~GDSCv2~\cite{gdsc}, 
    \textcolor[HTML]{4682B4}{\ding{110}}~DAVIS~\cite{davis}, 
    \textcolor[HTML]{8B4513}{\ding{110}}~KIBA~\cite{kiba}, 
    \textcolor[HTML]{FF7F50}{\ding{110}}~BindingDB~\cite{bindingdb}, 
    \textcolor[HTML]{4682B4}{\ding{108}}~DrugBank~\cite{drugbank}, 
    \textcolor[HTML]{FF7F50}{\ding{108}}~TWOSIDES~\cite{twosides}, 
    \textcolor[HTML]{2E8B57}{\ding{115}}~ZINC~\cite{zinc}, 
    \textcolor[HTML]{800080}{\ding{115}}~QM9~\cite{qm9-1,qm9-2}
    \textcolor[HTML]{8B4513}{\ding{108}}~USPTO~\cite{uspto}.\\

    \end{minipage}
    \caption{\textbf{Format and information of TCDD. a-d,} Sample structure with ShareGPT format for simple-turn dialogue. The dataset samples following the ShareGPT format comprise conversations, system, and tools.  \textbf{e,} Basic information of drug discovery tools including names, input/output parameter structures, data sources, and the specific quantities of samples in training set.}
    \label{fig:sharegpt}
\end{figure*}



\section*{Discussion}\label{sec12}

Recent advances in AI-assisted drug discovery have resulted in the development of transformative platforms such as Chemistry42 \cite{Chemistry42}, DrugFlow \cite{drugflow}, and large language model (LLM)-based agents, such as DrugAgent \cite{drugagent2,drugagent1}, DrugAssist \cite{ye2025drugassist}, and MolecularGPT \cite{liu2024moleculargpt}. However, three critical limitations persist: (1) non-expert researchers struggle with the technical demands of SOTA models \cite{sadybekov2023computational,zhavoronkov2019deep,talevi2023computer,morgnanesi2015computational}, (2) existing tools fail to align multimodal data, and (3) workflow fragmentation necessitates manual task integration \cite{zhang2025artificial,paul2010improve}. 

This paper proposes DrugPilot, an end-to-end  LLM-based parameterized reasoning agent for drug discovery to address these challenges. To represent multimodal data structurally and parametrically, as well as to break through the traditional text-based context length limitation, we innovatively propose PMP. The PMP solves the problem of multimodal data fragmentation by converting drug-related entities (e.g., SMILES expressions, protein sequences, etc.) into executable key-value pairs. The PMP solves the multimodal data problem by converting drug-related entities (e.g. SMILES expressions, protein sequences, etc.) into actionable key-value pairs. Unlike the text-based memory mechanism relied on by mainstream LLMs (e.g., ChatGPT, DeepSeek) and agents (e.g., CoT \cite{cot}, ReAct \cite{react}), PMP saves parameters structured in unstructured text as key-value pairs (Fig. \ref{fig:framework}b), which achieves lossless transfer of information and effectively improves the efficiency and accuracy of collaborative processing of multimodal data \cite{Multimodal}. Under the same task settings, DrugPilot keeps the task completion rate above 95\% when processing inputs of different scales (1$\sim$30 SMILES strings) compared to other agents (Fig.\ref{fig:ablation_mp}b). In addition, we also conducted a maximum parameter handling capacity test. The results show that compared with the GPT-4o \cite{gpt4o}, which supports 51 parameters at maximum, DrugPilot has a theoretically unlimited parameter processing capability (Fig.\ref{fig:ablation_mp}e).

To further improve the robustness of tool calling, we design the Fe-Fo mechanism. In the multiple-function category, adding only the Fe-Fo mechanism, DrugPilot improves the functional accuracy (Acc. F) and parameter accuracy (Acc. P) by 5.8\% and 13.7\%, respectively, which is significantly better than the ReAct (Table \ref{tab:model_performance}). Meanwhile, we constructed the TCDD, a drug discovery tool calling dataset for drug discovery tasks, which covers 8 core tasks with a total of 2,800 high-quality annotated samples, and supports SFT and evaluation. On the TCDD dataset, DrugPilot achieved superior task completion rates, reaching 98.0\% for simple function, 93.5\% for multiple function, and 64.0\% for multi-turn function, significantly outperforming all baseline methods in each category. Even without fine-tuning, DrugPilot significantly outperforms mainstream agents that have been fine-tuned (e.g., CoT \cite{cot}, ToT \cite{tot}, ReAct \cite{react}, etc.).

In summary, DrugPilot breaks down the key barriers between AI and the actual drug discovery process through a parametric reasoning structure, and a robust error correction mechanism. Its superior accuracy and efficient workflow automation in multi-tasking scenarios effectively alleviate the long-standing inefficiencies in the drug discovery process. While data construction and task extensibility remain future challenges, this study lays the groundwork for future research, such as incorporating reinforcement learning for dynamic task planning and enabling broader intelligent assists for drug discovery. DrugPilot demonstrates the transformative potential of LLM Agents in accelerating drug discovery by reducing technical barriers and enhancing cross-domain collaboration efficiency.

\section*{Methods}


To optimize the performance of LLM agents in executing drug discovery tasks through tool-calling, we design DrugPilot. First, we constructed the TCDD dataset and used it to fine-tune LLMs, aiming to enhance the models’ domain-specific knowledge in drug discovery and their understanding of drug-related tools. Second, we propose the parameterized memory pool mechanism to handle large-scale, multimodal data and multi-turn conversations. Finally, we propose a feedback-focus mechanism to help LLMs correct common errors when calling drug-related tools and to maintain focus on the original task.

The DrugPilot framework, depicted in Fig.\ref{fig:framework}a. The user’s drug discovery task is first provided as input to the LLM backbones. These LLMs will have the reason to make decisions on reading the PMP and calling tools. After verification by the Fe-Fo mechanism, the tool calls are executed, and results are returned. If errors are detected during verification, feedback is provided to the LLMs through the Fe-Fo mechanism. Throughout this process, users can view or update parameters in the PMP at any time to oversee the task execution. The LLMs will continuously reason based on the above information until the final answer is produced.

\subsection*{Tool-Calling Dataset for Drug Discovery}

LLMs have achieved strong performance in general tool calling tasks, but given the lack of knowledge in relevant fields \cite{holstein2024bridging}, complex data forms, and the need for batch processing of data \cite{zhu2020big}, LLMs still struggle to correctly infer tool names and parameters of drug discovery tasks. To address these challenges, we built TCDD, a tool-calling dataset for drug discovery to fine-tune and test LLMs. Specifically, TCDD is used to validate the three key capabilities of LLMs:
\begin{enumerate}
    \item Tool selection: comprehending natural language queries of drug discovery tasks and selecting corresponding drug discovery tools;
    \item Parameter extraction: identifying and extracting required parameters of unique data forms from the context.
    \item Interaction with PMP: retrieving parameterized data in PMP and saving execution results.
\end{enumerate}

The TCDD simulates real-world conversations between users and AI systems in the area of drug discovery, comprising various scenarios such as single-turn, multi-turn, error correction, and memory pool updates. It has a total of 2,800 instruction samples, 2500 for training and 300 for testing, with the ShareGPT format, which is originally designed for multi-turn dialogue. 
This format supports LLMs in interacting with external services and thus is particularly suitable for drug discovery tasks characterized by multi-turn workflows and tool calling. Fig.~\ref{fig:sharegpt}e shows the basic information of the eight drug discovery tools in the TCDD, containing their names, input/output parameter structures, data sources, and the specific quantities of samples for different dialogue patterns in the training set.

Since the interaction with the memory pool requires formalized parameters, the samples are categorized into two types based on whether the memory pool mechanism is employed. Fig.~\ref{fig:sharegpt}a and Fig.~\ref{fig:sharegpt}b illustrates examples of both types of dialogues using the \texttt{drug\_property} tool to predict the aqueous solubility of a given drug. Each sample consists of three components: conversation, system instructions, and tools, with variations in the conversation portion between the two sample types. The conversation segment includes:
\begin{enumerate}
    \item Human: The user inputs a natural language description of the drug discovery task;
    \item Function Call: The LLM generates the tool calling information in JSON format, specifying the tool name and corresponding parameters;
    \item Observation: The tool is called and the execution result is returned;
    \item LLM: The LLM generates a response integrating the observation and task objectives.
\end{enumerate}
The system section guides the LLM to execute tool calling and return results in a standardized format. The tools section provides essential information about the available tools, including: function names and descriptions, acceptable parameters with types and descriptions, and required parameters.

During the development of TCDD, we designed single-turn/multi-turn dialogue patterns, diversified user query expressions, and varied drug molecule types to enhance performance. Approximately 50\% of the samples consist of single-turn dialogues with complete instructions, establishing fundamental tool calling rules and memory pool retrieval mechanisms. The remaining samples simulate complex multi-turn dialogues reflecting real-world workflows, including multi-turn tasks (30\%) and parameter error scenarios (20\%). For instance, in a typical drug optimization workflow, the system first generates molecular candidates based on inhibitory concentrations against specific cell types, then predicts drug properties (e.g., solubility), evaluates drug-target binding affinity, and finally refines the candidates. Such workflows require the model to integrate intermediate results from prior steps, efficiently extract parameters using the memory pool, and handle error propagation when tool failures occur. For common user input issues such as missing or misspelled parameters, the tool calling returns observations containing specific error messages. Corresponding samples in the dataset train the LLM to accurately discern user intent and interact appropriately based on tool specifications and observations.

The distribution of samples across different tools and patterns was determined considering their usage scenarios, parameter complexity, and frequency of application. In a complete workflow, tools such as \texttt{drug\_optimization} and \texttt{drug\_generation} are more likely to be used in combination with others. Hence, their ratios of multi-turn to single-turn samples are relatively higher. Since \texttt{drug\_target\_affinity} and \texttt{drug\_target\_interaction} require target protein sequences as parameters, which are typically long and complex, these tools have more error-correction samples. Similarly, given that \texttt{drug\_optimization} involves significantly more parameters than other tools, it also includes a higher proportion of error-handling samples.




\begin{figure*}[t!]
    \centering
     \includegraphics[width=0.95\linewidth]{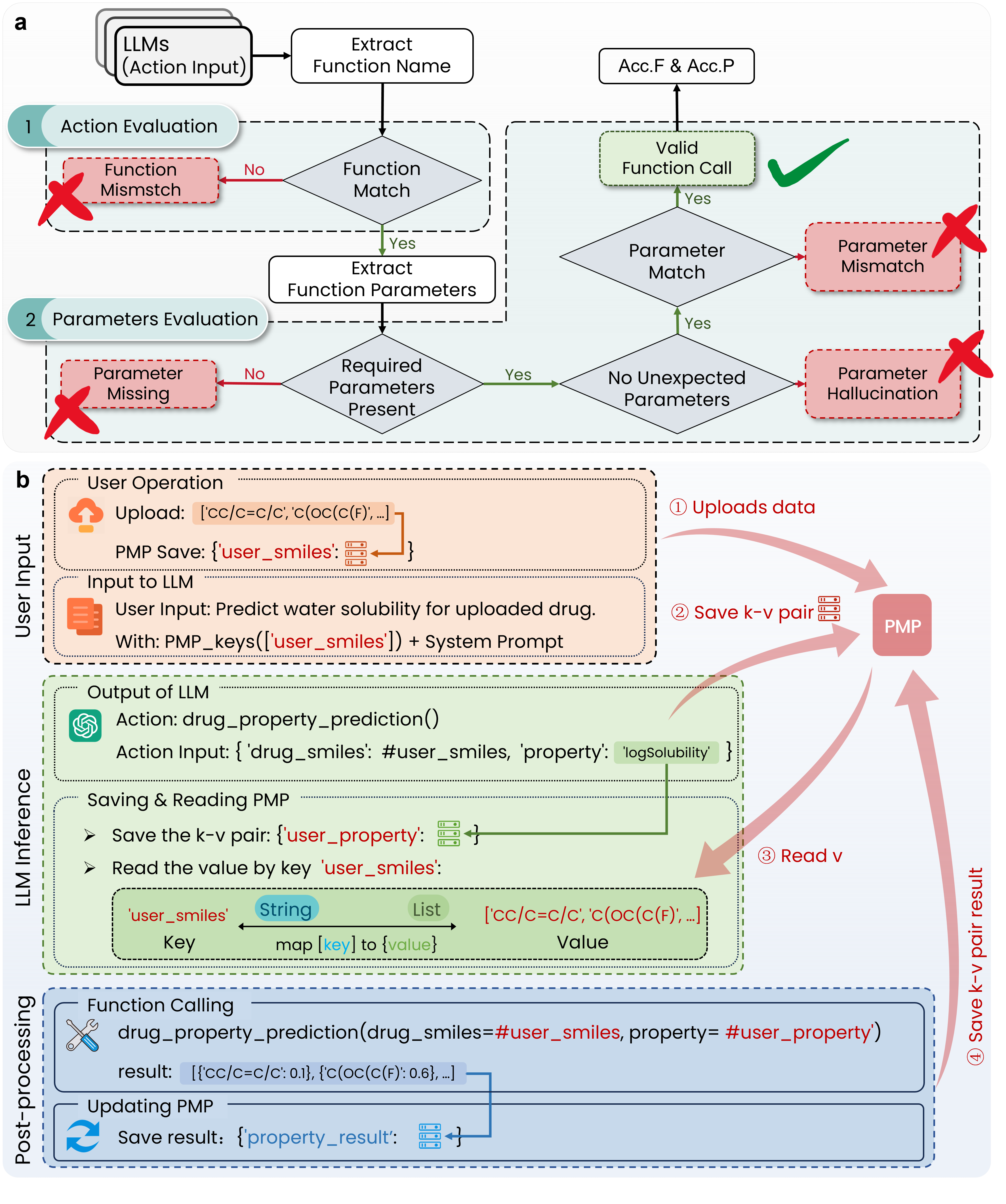}
    \caption{\textbf{Process of evalution and parameter management in DrugPilot.} \textbf{a,} The step by step evaluation process. \textbf{b,} An example of PMP. PMP automatically saves parameters from conversations and performs key-value mapping, allowing users to modify them anytime. }
    \label{fig:evaluationandpmpcase}
\end{figure*}

\subsection*{Parameterized Memory Pool}


To address the limitations of traditional memory modules in transmitting large-scale, multimodal drug-related parameters, we propose a novel memory module: the parameterized memory pool (PMP). PMP does not store unstructured text but instead maintains structured key-value pairs. Its purpose is to optimize parameter passing efficiency within the agent, thereby reducing the reasoning burden on LLMs, improving the controllability of reasoning results, and ultimately enhancing the performance of LLMs in drug-related tool calling. We conducted prompt engineering to help LLMs better understand PMP. The memory pool prompt explicitly defines the purpose, usage scenarios, and usage method of PMP, as detailed in Appendix~\ref{app:mp_prompt}.

\paragraph{The Structure of PMP}

In traditional memory modules, memory takes the form of a series of conversation text fragments, with drug-related parameters embedded within the text. While the PMP is a key-value store, where each key-value pair maintains a parameter related to drug discovery. Each key is unique and is a short string representing the name of a drug-related parameter, designed to convey its meaning as clearly as possible for interaction with LLMs. The value is the actual content of the parameter, structured in a format that can be directly used as input for drug-related tools, enabling seamless interaction between PMP and external tools. When there are multiple parameters of the same type, they share one key, and their corresponding values are stored as a list containing all the parameters of that type. 

As shown in Fig.\ref{fig:framework}b, when storing a large-scale list of drug molecules, the key could be \texttt{"generated\_drug\_smiles"}, indicating that it represents a user-provided list of molecular expressions. Such drug-related parameters can be of considerable scale, for example, they may consist of lists with tens of thousands of entries and be embedded within complex textual contexts. This large-scale, multimodal context poses significant challenges for both storage and reasoning. The PMP extracts these large-scale parameters from complex text and converts them into concise keys, allowing LLMs to interact only with these short keys. This greatly reduces the length of context required for the task. Furthermore, PMP maintains the parameters in a structured manner, enabling subsequent tool invocations to directly use these parameters without requiring LLMs to repeatedly infer their complete content.

\paragraph{Parameter Reading and Updating}

In traditional memory modules, the stored memory content is typically text fragments from the conversation. When performing reasoning with the LLMs, the user inputs a piece of text, and memory modules will also retrieve a piece of text from the stored historical information. These two parts of the text are concatenated and fed into the LLMs together for reasoning. This parameter extraction process can be formulated as:
\begin{equation}
v = \mathcal{R}_{LLM}(T_u + T_m), \quad \text{where } T_m \subseteq \mathcal{M}
\end{equation}
where, $\mathcal{M}$ denotes the entire stored memory, $T_m$ denotes the retrieved memory text,$T_u$ denotes the user input text, $\mathcal{R}_{LLM}$ denotes the function for retrieving parameters from text using LLMs and $v$ denotes the final parameter passed to the tool. In this method, the large-scale complex texts bring a tremendous inference burden for LLMs

Fig.\ref{fig:evaluationandpmpcase}b illustrates the workflow of the PMP. First, before the conversation begins, users can upload their own datasets or public datasets into DrugPilot. It is important to note that users are not limited to uploading parameters only at the beginning. They can add, delete, modify, and query parameters in the PMP at any time. This ability to dynamically control and adjust the task direction at any stage enhances the flexibility of the memory module. Compared to disorganized large blocks of text, the structured memory format enables users to operate on memory content. This feature allows humans to efficiently interact with the memory of large models and significantly improves the flexibility of the memory module.

Next, after the drug discovery task and the current keys stored in PMP are input into the LLM, the LLM analyzes the input text, selects the tool to call, and considers how to obtain the parameters required by that tool. If the required parameters are included in the input text, the LLM will identify and extract them directly from the text. Meanwhile, PMP saves the parameter as a key-value pair by assigning a key to the parameter and storing its content in the corresponding value for use in the next step. If the key already exists, PMP will use a list as the value for that key and append the current parameter to the list.

If the user's input text does not contain the required parameter, the LLM will select a key corresponding to the required parameter from the list of keys in the PMP. The PMP will then map the selected key to its corresponding value and pass that value as a parameter to the tool. If the value is a list containing multiple parameters, the PMP will take the last element of the list, which is the most recently added parameter. Through the key-value conversion mechanism of PMP, LLMs' reasoning task becomes selecting the key to use. It makes LLMs no longer interact directly with large-scale data, effectively reducing the burden on various components of the LLM-based agent system, including LLM inference, memory storage, and memory retrieval. The parameter extraction process of DruPilot can be formulated as:
\begin{align}
    v &= \mathcal{G} (\underset{k \in \mathcal{M_K}}{\arg\max}\ \mathcal{P}_{LLM}(\mathcal{M_K}))
\end{align}
where, $\mathcal{M_K}$ denotes the key set in the memory pool, $k$ denotes a key from $\mathcal{M_K}$, $\mathcal{P}_{LLM}$ denotes the function using LLMs to get the probability of using $k$, and $\mathcal{G}$ denotes the predefined mapping function. The PMP provides the LLMs with a series of selectable parameter keys. 

After a successful tool calling, PMP will save the result returned by the tool as a key-value pair. It assigns a key to the result and stores the entire result as the value. In the next step, this result will be added to PMP's key list and passed as input to the LLM. At this point, the LLM can choose this execution result as the input parameter for the next tool invocation.

\subsection*{Feedback-focus Mechanism}

LLMs cannot always perfectly select tools and pass parameters with complete correctness in both content and format. Additionally, LLMs tend to forget the initial task during long conversations. To address these issues, we propose a feedback-focus mechanism, named Fe-Fo, to help LLMs correct errors and maintain focus, making their output more controllable.

In the feedback mechanism, we alleviate common reasoning issues by feeding error information back to LLMs. When calling drug-related tools, the reasoning output of LLMs exhibits a series of common issues, as illustrated in Appendix \ref{app:reason_errors}. We have verified these issues and designed corresponding feedback prompts for each error type. These prompts describe the type and cause of the error in detail and explicitly instruct the LLMs to regenerate the output according to the requirements based on this information.

In the focus mechanism, we help LLMs stay focused by reiterating the original task. If LLMs are unable to resolve the issue in one attempt, the length of the conversation will inevitably increase. At this point, LLMs often forget the initial task, leading to a loss of focus and resulting in aimless attempts, such as randomly changing the selected tool or passing parameters based on hallucinations. To address this issue, when LLMs make a reasoning error, DrugPilot will repeat the original task to the LLMs, ensuring that they remain focused on the required drug discovery task.

Fe-Fo integrates the above two types of information and provides a clear instruction: regenerate the output according to the formatting requirements based on the given information. When any error from the types shown in Appendix \ref{app:reason_errors} is detected, the Fe-Fo mechanism will feed the Fe-Fo prompt into the LLMs, which can be formulated as:
\begin{align}
    O_{t+1} &= LLM(\mathcal{E}(O_t) + T + I)
\end{align}
where, ${I}$ denotes the instruction to guide the reasoning, ${T}$ denotes the original task description, $\mathcal{E}$ denotes the error detection function applied to the output of LLMs which returns the feedback information corresponding to the error and ${O_t}$ denotes the output of LLMs at step $t$.

\subsection*{Evaluation Metrics}

Drawing on the Berkeley function-calling leaderboard \cite{gorilla}, we designed a custom evaluation framework for DrugPilot and other agents. A valid tool calling requires correct tool selection, accurate parameter extraction, and effective self-correction across multiple turns. Therefore, we divide the evaluation into three categories, with 100 queries in each category, which together constitute the test set.
\begin{itemize}
    \item Simple function: This category contains the simplest situation with one and only one function supplied.
    \item Multiple function: This category contains a user query that requires the invocation of only one function among eight available tools.
    \item Multi-turn function: This category contains multi-turn queries, where different queries may correspond to different tool callings. 
\end{itemize}

\paragraph{Baselines}  DrugPilot integrates multiple drug discovery tools, and it mainly focuses on accurate tool calling to complete a full workflow of drug discovery tasks. However, some existing methods, like DrugAgent \cite{drugagent1}, are dedicated to improving performance on individual tasks, concentrating on text understanding and result interpretation. These methods do not align well with our task objectives.
Therefore, we select four representative agents with tool-calling capabilities in recent times for comparison, and these four baseline agents are based on pre-trained LLMs: 
\begin{itemize}
    \item CoT \cite{cot} is a simple baseline where the agent generates a solution by breaking the problem into substeps.
    \item LoT \cite{lot} is a self-improvement prompting framework where the agent verifies and refines its intermediate reasoning steps by embedding symbolic logic principles.
    \item ReAct \cite{react} interleaves reasoning traces with actions to enable LLMs to iteratively plan, gather information, and adjust strategies. 
    \item ChatGPT-4o \cite{gpt4o} Function Calling API is a structured tool-calling interface provided by OpenAI, built on the closed-source LLM and tool-calling method.
\end{itemize}

\paragraph{Fine-Tuning} We fine-tuned a series of small-scale LLMs on TCDD using LoRA \cite{lora}, including Meta-Llama-3.1-8B-Instruct \cite{llama3_1}, Meta-Llama-3-8B-Instruct \cite{llama3}, Mistral-Nemo-Instruct-2407 \cite{mistral}, Gemma-2-9B-it \cite{team2024gemma}, Qwen2-7B-Instruct \cite{qwen2}, DeepSeek-LLM-7B-Chat \cite{deepseek-llm},  DeepSeek-R1-Distill-Llama-8B \cite{deepseek-r1} and Llama-3-Groq-8B-Tool-Use (\url{https://groq.com/introducing-llama-3-groq-tool-use-models.}). We use the open-source dataset glaive\_toolcall (\url{https://huggingface.co/datasets/glaiveai/glaive-function-calling-v2.}) to improve the general tool-calling capabilities of the LLMs. And we divided our TCDD into training, validation, and testing subsets at a ratio of 8:1:1 to enhance LLMs' specialized capabilities in drug-related tool calling and evaluated their performance. Details of the hyperparameter settings employed during fine-tuning is recorded in Appendix~\ref{app:fine-tuning}.

\paragraph{Evaluation Process}
For each task, the LLM generates an action input, information required to call the tools, in JSON format. Given that hallucination remains a challenge for LLMs, especially in tool calling, we evaluate their responses in two aspects: tool selection and parameter extraction. According to this, we report two accuracy metrics: function accuracy (\textbf{Acc.F}) and parameter accuracy (\textbf{Acc.P}). Additionally, detailed calculation formulas are shown in Appendix \ref{app:acc_formula}.

Fig.\ref{fig:evaluationandpmpcase}a shows the step-by-step evaluation process of the LLM response, namely the action input. First, the action input undergoes function evaluation, where the tool name is extracted and verified. Once the tool name is correct, the function parameters are parsed and undergo parameter evaluation. This step verifies whether all required parameters are present and that no unexpected parameter is included. Finally, each parameter value is inspected to confirm its correctness and integrity. Only when all these checks are completed is the tool calling deemed valid. The two parts of the evaluation correspond respectively to function accuracy and parameter accuracy in the final results.  In addition, the maximum allowable execution time for an individual query is set to 120s; any response that exceeds this threshold is directly determined erroneous.

\section*{Data availability}
The tool-calling dataset for drug discovery (TCDD), which includes data for both fine-tuning and evaluation, is available at \url{https://drive.google.com/file/d/1JthOkIAzuuaajZhgH03e9TfM9KwBHmni/view?usp=sharing.}

\section*{Code availability}
The source code of DrugPilot is freely available and can be found on the GitHub at \url{https://github.com/wzn99/DrugPilot}.

\begin{table}[t!]
\centering
\setlength{\tabcolsep}{2mm}
\resizebox{\textwidth}{!}{%
\begin{tabular}{ccccccccccccc}
\toprule
\multicolumn{1}{c}{} &\multicolumn{1}{c}{} & \multicolumn{1}{c}{} &\multicolumn{2}{c}{DrugPilot (ours)}&\multicolumn{2}{c}{DrugPilot w.o. SFT}& \multicolumn{2}{c}{CoT}& \multicolumn{2}{c}{LoT} &  \multicolumn{2}{c}{ReAct}\\ \cmidrule(lr){4-13}
\multicolumn{1}{c}{\multirow{-2}{*}{Category}}&\multicolumn{1}{c}{\multirow{-2}{*}{Model}}&\multicolumn{1}{c}{\multirow{-2}{*}{Scale}} 
& \makebox[0.05\textwidth][c]{Acc.F\textit{\textsuperscript{b}}}   &\makebox[0.05\textwidth][c]{Acc.P\textit{\textsuperscript{b}}} & \makebox[0.05\textwidth][c]{Acc.F}   &\makebox[0.05\textwidth][c]{Acc.P}& \makebox[0.05\textwidth][c]{Acc.F}   &\makebox[0.05\textwidth][c]{Acc.P}& \makebox[0.05\textwidth][c]{Acc.F}   &\makebox[0.05\textwidth][c]{Acc.P}& \makebox[0.05\textwidth][c]{Acc.F}   &\makebox[0.05\textwidth][c]{Acc.P}\\ \midrule

\multicolumn{1}{c}{\multirow{8}{*}{\rotatebox{90}{Simple Function}}} 
& Llama3.1&8B 
&$\textbf{98.4}_{~1.2}$\textit{\textsuperscript{cd}}&$ \textbf{98.0}_{~1.3} $
&$89.4_{~0.4}$&\underline{$88.4_{~1.0}$}\textit{\textsuperscript{d}}
&\underline{$92.6_{~2.9}$}&$63.4_{~3.4}$
&$81.2_{~5.1}$&$46.4_{~4.6}$
&$86.6_{~4.4}$&$86.6_{~4.4}$\\
&Llama3&8B 
&$\textbf{99.2}_{~0.8}$&$\textbf{97.6}_{~2.1}$
&$73.0_{~3.9}$&$73.0_{~3.9}$
&\underline{$81.0_{~3.5}$}&\underline{$75.0_{~3.2}$}
&$43.4_{~4.5}$&$38.8_{~5.3}$
&$71.0_{~4.2}$&$71.0_{~4.2}$
\\
&Mistral-NeMo&7B 
&$\textbf{97.4}_{~0.5}$&$\textbf{97.0}_{~1.0}$
&$89.2_{~1.8}$&$89.2_{~1.8}$
&$90.8_{~4.9}$&$89.0_{~5.4}$
&\underline{$93.4_{~2.8}$}&\underline{$89.4_{~4.2}$}
&$88.6_{~4.8}$&$88.6_{~4.8}$\\
&Gemma2&9B 
&$\textbf{97.8}_{~1.1}$&$\textbf{97.4}_{~1.1}$
&\underline{$90.2_{~0.1}$}&\underline{$90.2_{~0.1}$}
&$87.0_{~0.1}$&$86.4_{~1.1}$
&$85.8_{~1.8}$&$80.6_{~1.1}$
&$88.0_{~0.1}$&$87.6_{~0.5}$\\
&Qwen2&7B 
&$\textbf{99.3}_{~0.1}$&$\textbf{95.9}_{~1.8}$
&\underline{$98.6_{~0.1}$}&\underline{$89.2_{~3.9}$}
&$98.2_{~1.1} $&$87.8_{~1.5}$
&$97.0_{~1.6}$&$83.2_{~3.1}$
&$97.4_{~0.5}$&$88.4_{~2.3}$\\
&DeepSeek-LLM&7B 
&$\textbf{47.6}_{~2.6}$&$ \textbf{41.2}_{~4.0}$
&$33.9_{~4.7}$&\underline{$33.6_{~4.1}$}
&$17.2_{~4.7}$&$13.6_{~2.7}$
&$19.6_{~3.6}$&$11.2_{~4.0}$
&\underline{$40.4_{~3.4}$}&$27.4_{~3.6}$\\
&DeepSeek-R1\textit{\textsuperscript{a}}&8B 
&$\textbf{97.2}_{~0.8}$&$\textbf{73.6}_{~6.2}$
&\underline{$66.2_{~0.1}$}&\underline{$61.8_{~1.3}$}
&$48.2_{~5.4}$&$43.6_{~4.2}$
&$61.0_{~3.3}$&$54.8_{~2.6}$
&$58.0_{~4.6}$&$56.6_{~4.3}$\\
&Llama3-Groq&8B
&$\textbf{99.8}_{~0.4}$&$\textbf{90.6}_{~2.4}$
&\underline{$44.0_{~5.6}$}&\underline{$43.0_{~5.4}$}
&$15.8_{~1.9}$&$15.2_{~2.3}$
&$23.2_{~2.9}$&$22.6_{~3.2}$
&$38.6_{~4.5}$&$38.6_{~4.5}$\\ \hline

\multicolumn{1}{c}{\multirow{8}{*}{\rotatebox{90}{Multiple Function}}}
&Llama3.1&8B 
&$ \textbf{98.7}_{~0.6}$&$ \textbf{93.5}_{~2.3}$
&\underline{$79.4_{~2.9}$}&\underline{$58.2_{~1.2}$}
&$55.8_{~3.3}$&$50.5_{~3.4}$
&$53.5_{~2.7}$&$43.0_{~3.6}$
&$76.3_{~5.1}$&$56.3_{~5.6}$\\
&Llama3&8B 
&$\textbf{98.0}_{~2.3}$&$\textbf{92.8}_{~4.6}$
&\underline{$65.0_{~1.7}$}&\underline{$54.3_{~2.0}$}
&$46.3_{~5.4}$&$43.3_{~5.6}$
&$35.8_{~4.6}$&$32.8_{~4.2}$
&$59.8_{~2.1}$&$53.5_{~2.7}$\\
&Mistral-NeMo&7B 
&$ \textbf{98.4}_{~2.1}$&$ \textbf{96.3}_{~2.5}$
&$86.8_{~4.8}$&\underline{$83.7_{~3.8}$}
&$82.0_{~2.7}$&$68.3_{~3.4}$
&$76.0_{~3.8} $&$70.8_{~3.0}$
&\underline{$92.3_{~3.5}$}&$77.8_{~3.4}$\\
&Gemma2&9B 
&$ \textbf{99.5}_{~0.7} $&$ \textbf{92.2}_{~2.4}$
&$91.0_{~2.2}$&$72.4_{~3.1}$
&$92.5_{~2.7}$&\underline{$79.8_{~1.9}$}
&$86.8_{~2.9}$&$76.3_{~4.2}$
&\underline{$93.0_{~1.4}$}&$73.8_{~2.3}$\\
&Qwen2&7B 
&$ \textbf{93.7}_{~2.9} $&$ \textbf{87.2}_{~2.3}$
&\underline{$93.2_{~2.9}$}&$76.0_{~5.7}$
&$87.5_{~2.5}$&\underline{$77.8_{~4.5}$}
&$83.3_{~4.6}$&$71.5_{~3.0}$
&$92.5_{~1.1}$&$74.5_{~1.9}$\\
&DeepSeek-LLM&7B 
&$\textbf{39.5}_{~2.7} $&$\textbf{18.8}_{~1.6}$
&\underline{$26.5_{~6.5}$}&\underline{$13.0_{~2.1}$}
&$11.0_{~2.1}$&$\ \ 5.3_{~2.1}$
&$\ \ 9.5_{~3.3} $&$\ \ 5.0_{~3.5}$
&$20.5_{~2.4}$&$\ \ 8.5_{~2.2}$\\
&DeepSeek-R1&8B 
&$ \textbf{93.9}_{~2.0} $&$ \textbf{72.2}_{~4.1}  $
&\underline{$60.3_{~5.3}$}&\underline{$49.0_{~2.2}$}
&$51.8_{~3.4}$&$34.3_{~3.3}$
&$39.0_{~3.6}$&$29.0_{~3.8}$
&$60.3_{~4.8}$&$43.3_{~4.0}$\\
&Llama3-Groq&8B 
&$\textbf{96.2}_{~0.9} $&$ \textbf{78.5}_{~2.1} $
&$39.5_{~4.2 }$&\underline{$30.5_{~3.2}$}
&$17.5_{~3.5}$&$17.0_{~3.1}$
&$10.3_{~1.9}$&$\ \ 9.3_{~1.4}$
&\underline{$40.3_{~7.4}$}&$28.8_{~4.6}$\\ \hline

\multicolumn{1}{c}{\multirow{8}{*}{\rotatebox{90}{Multi-turn Function\textit{\textsuperscript{e}}}}}
&Llama3.1&8B
&$ \textbf{72.7}_{~6.7} $&$ \textbf{64.0}_{~5.7} $
&\underline{$49.5_{~2.4}$}&\underline{$38.2_{~2.3}$}
&$43.0_{~3.6}$&$31.8_{~2.9}$
&$30.1_{~5.4}$&$24.5_{~4.8}$
&$45.1_{~5.2}$&$35.5_{~3.7}$\\
&Llama3&8B
&$\textbf{74.2}_{~4.1} $&$\textbf{61.9}_{~4.7} $
&\underline{$34.8_{~4.7}$}&\underline{$25.1_{~5.4}$}
&$25.8_{~5.9}$&$17.3_{~3.8}$
&$20.7_{~3.3}$&$14.7_{~2.6}$
&$32.6_{~1.4}$&$18.0_{~2.6}$\\
&Mistral-NeMo&7B
&$ \textbf{70.0}_{~6.1} $&$ \textbf{56.9}_{~4.1}$
&\underline{$54.9_{~5.1}$}&$37.2_{~2.9}$
&$44.4_{~3.2}$&$33.2_{~3.1}$
&$38.9_{~3.8}$&$31.3_{~1.6}$
&$50.9_{~2.9}$&\underline{$39.5_{~3.0}$}\\
&Gemma2&9B
&$ \textbf{84.9}_{~1.9}$&$ \textbf{61.8}_{~4.1} $
&$44.3_{~1.8}$&$34.4_{~2.1}$
&\underline{$54.3_{~2.3}$}&\underline{$ 43.1_{~2.6}$}
&$35.3_{~3.4}$&$27.6_{~3.1}$
&$34.5_{~2.1}$&$32.8_{~1.9}$\\
&Qwen2&7B
&$ \textbf{73.9}_{~3.2}$&$ \textbf{57.3}_{~3.3}$
&\underline{$38.2_{~4.0}$}&\underline{$22.7_{~2.6}$}
&$34.2_{~3.9}$&$20.1_{~2.6}$
&$35.2_{~3.8}$&$20.2_{~3.9}$
&$21.4_{~1.1}$&$18.7_{~1.2}$\\
&DeepSeek-LLM&7B
&$\textbf{19.9}_{~4.5}$&$\ \ \textbf{9.0}_{~2.3}$
&$\ \ 4.1_{~1.0}$&\ \ \underline{$2.4_{~0.3}$}
&$\ \ 1.8_{~0.8}$&$\ \ 0.4_{~0.5}$
&$\ \ 1.7_{~0.5}$&$\ \ 0.5_{~0.7}$
&\ \ \underline{$9.7_{~2.0}$}&$\ \ 1.9_{~0.3}$\\
&DeepSeek-R1&8B
&$ \textbf{71.5}_{~1.8} $&$ \textbf{51.4}_{~3.8} $
&$25.6_{~2.3}$&\underline{$19.1_{~2.5}$}
&\underline{$27.4_{~4.2}$}&$17.9_{~2.2}$
&$\ \ 7.4_{~1.9}$&$\ \ 6.8_{~2.3}$
&$16.3_{~3.9}$&$15.2_{~4.3}$\\
&Llama3-Groq&8B
&$\textbf{79.6}_{~3.4} $&$ \textbf{49.0}_{~5.1}$
&\underline{$24.4_{~1.9}$}&\underline{$19.4_{~2.2}$}
&$\ \ 8.9_{~1.2}$&$\ \ 7.1_{~1.0}$
&$\ \ 8.5_{~2.5}$&$\ \ 5.6_{~1.5}$
&$12.4_{~3.0}$&$11.8_{~3.4}$\\ \hline
\end{tabular}
}
\begin{minipage}{\textwidth} 
\tiny
\footnotetext[a]{The model Deepseek-R1 with 8B parameters stands for DeepSeek-R1-Distill-Llama-8B.}
\footnotetext[b]{The two accuracy rates, Acc.F and Acc.P are the average results of repeat evaluation experiments with the standard deviation attached as a corner mark. }
\footnotetext[c]{The subscript of each accuracy rate indicates the standard deviation over five measurements. }
\footnotetext[d]{\textbf{Bolded} entries denote the highest accuracy rates, and \underline{underlined} entries denote the second-highest.}
\footnotetext[e]{In the multi-turn category, more weight is put on the later queries and thus its accuracy, or score, is more suitable, reflecting more authentic performance on multi-stage tasks.}
\end{minipage}
\caption{\textbf{Overall results of experiments.} The tool calling performance on three categories of different LLMs and agents are measured by two accuracy metrics. CoT, LoT, and ReAct are baseline methods, and DrugPilot is combined with SFT and pre-trained LLMs.}
\label{tab:model_performance}
\end{table}

\newpage
\begin{appendices}




\section{Memory Pool Prompt}
\label{app:mp_prompt}

\begin{figure*}[ht!]
    \centering
    \includegraphics[width=\linewidth]{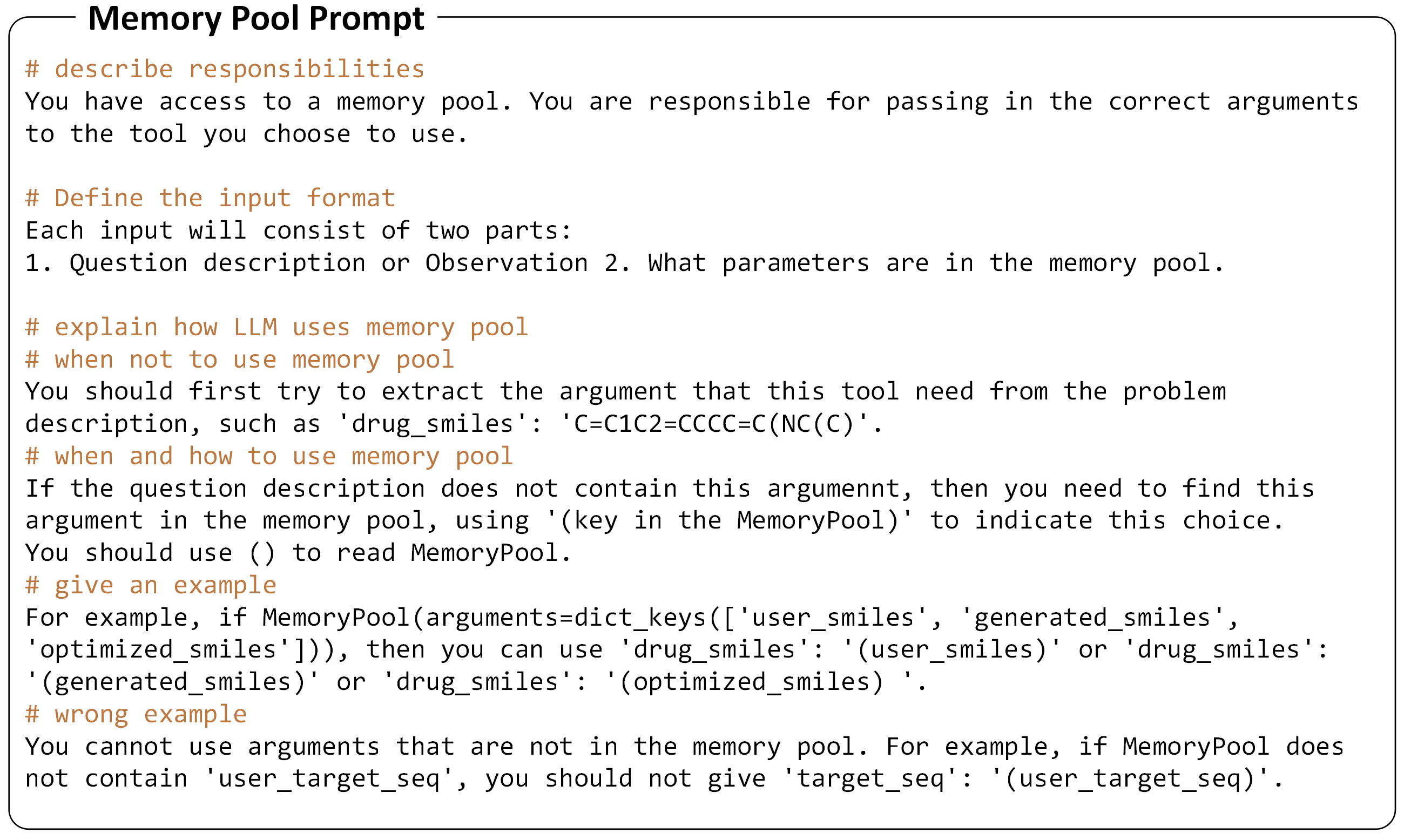}
    \caption{Memory pool prompt.}
    \label{fig:mp_prompt}
\end{figure*}

To help LLMs better understand the PMP in DrugPilot, we have incorporated a memory pool prompt into the system prompt. The full memory pool prompt is shown in Fig.~\ref{fig:mp_prompt}.

The memory pool prompt first clarifies the existence of PMP and the responsibilities of the LLMs, namely the correct transmission of parameters to the tools. It then defines the input format received by the LLMs, comprising two parts: the user’s question or the tool’s output, and a description of the current state of PMP, which includes the list of currently stored keys. LLMs can select a key from this pool and map it to its corresponding value. It then explains in detail how the LLMs should interact with PMP. First, it defines scenarios where PMP should not be used: if the required parameters are already present in the question, the LLMs should extract them directly. Next, it specifies when and how to use PMP: if the question lacks the necessary parameters, the LLMs must retrieve the corresponding key from the memory pool and enclose it in parentheses to indicate retrieval. Finally, the prompt provides both a correct and an incorrect example, demonstrating proper memory pool usage and helping LLMs avoid retrieving non-existent keys, thereby mitigating hallucination.

\section{DrugPilot's Reasoning Errors}
\label{app:reason_errors}

\begin{figure*}[h!]
    \centering
    \includegraphics[width=0.75\linewidth]{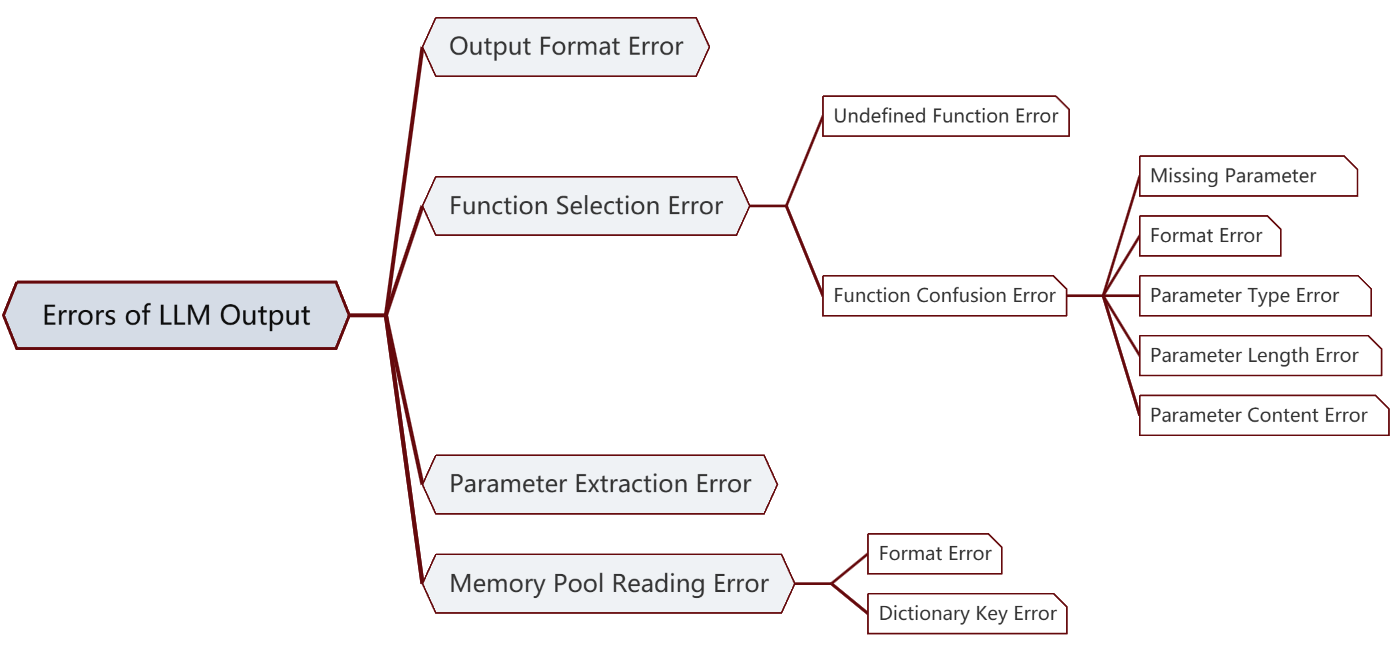}
    \caption{Common reasoning errors of LLMs. The common types of reasoning errors when LLMs call drug-related tools, and the Fe-Fo mechanism will provide feedback to LLMs regarding these issues.}
    \label{fig:reason_errors}
\end{figure*}

In tool calling, LLMs are required to generate an action input in JSON format, containing the tool name to be called and required parameters. And in actual tasks, there will be frequent interactions with PMP. Therefore, problems will inevitably arise both in content and format. Based on the real output of LLMs, we summarized the common reasoning errors as shown in Fig.\ref{fig:reason_errors}.

\section{Fine-Tuning Configuration}
\label{app:fine-tuning}
We conducted LoRA fine-tuning on the LLMs used in DrugPilot to enhance their domain knowledge in drug discovery and improve their ability to call drug-related tools. Batch size of 4 was used for smaller models, and 8 for larger ones. We deployed the final inference-stage LLMs on the Ollama\footnote{\url{https://github.com/ollama/ollama}.} platform. The hyperparameter settings used during the fine-tuning process are detailed in Table \ref{tab:hyperparams}.

\begin{table}[ht]
\centering
\resizebox{0.6\textwidth}{!}{%
\begin{tabular}{ll}
\toprule
\textbf{Hyperparameter} & \textbf{Value / Strategy} \\
\midrule
Batch size & 4-8 \\
Cutoff length & 1024 \\
Optimizer & AdamW \\
Initial learning rate & 5e-5 \\
Learning rate scheduler & Cosine decay \\
Precision & BF16 \\
Number of epochs & 3 \\
Deployment platform & Ollama \\
\bottomrule
\end{tabular}
}
\caption{Hyperparameter Settings for Fine-tuning}
\label{tab:hyperparams}
\end{table}

\section{Accuracy Calculation}
\label{app:acc_formula}
Acc.F and Acc.P represent the accuracy of tool selection and parameter extraction, and they are defined by:
\begin{align*}
    Acc.F = \frac{1}{N_{c}}\sum_{i=1}^{N_{c}}F(sample_i) \\
    Acc.P = \frac{1}{N_{c}}\sum_{i=1}^{N_{c}}P(sample_i)
\end{align*}
where $N_c$ is the number of samples in a category, $F(\cdot)$ and $P(\cdot)$ are indicator functions denote whether the current sample has correctly selected the function and extracted the parameters.





\end{appendices}


\end{document}